\documentclass{article}

\PassOptionsToPackage{numbers, compress}{natbib}
\usepackage[preprint]{neurips}

\usepackage[utf8]{inputenc} 
\usepackage[T1]{fontenc}    
\usepackage{hyperref}       
\usepackage{url}            
\usepackage{booktabs}       
\usepackage{amsfonts}       
\usepackage{nicefrac}       
\usepackage{microtype}      
\usepackage{xcolor}         
\usepackage{bbding}
\usepackage{wrapfig}
\usepackage{graphicx}
\usepackage{upquote}
\usepackage{subcaption}
\usepackage{multirow}

\usepackage{wrapfig}

\usepackage{textcomp}  
\usepackage{scalerel}  
\usepackage{xspace}
\usepackage{amsmath}

\newcommand{\modelname}{MARIO\xspace}

\usepackage{booktabs}
\usepackage{pgfplots}

\usepackage{algpseudocodex}
\usepackage{algorithm}
\usepackage{tikz}

\algnewcommand\algorithmictry{\textbf{try}}%
\algnewcommand\algorithmicexcept{\textbf{except}}%

\makeatletter
\algdef{SE}[TRY]{Try}{EndTry}[1]{\algpx@startIndent\algpx@startCodeCommand\algorithmictry\ #1}{\algpx@endIndent\algpx@startCodeCommand\algorithmicend\ \algorithmictry}%

\algdef{SE}[EXCEPT]{Except}{EndExcept}[1]{\algpx@startIndent\algpx@startCodeCommand\algorithmicexcept\ #1}{\algpx@endIndent\algpx@startCodeCommand\algorithmicend\ \algorithmicexcept}%

\ifbool{algpx@noEnd}{%
    \algtext*{EndTry}%
    \algtext*{EndExcept}%
    \apptocmd{\EndTry}{\algpx@endIndent}{}{}%
    \apptocmd{\EndExcept}{\algpx@endIndent}{}{}%
}{}%

\pretocmd{\Try}{\algpx@endCodeCommand}{}{}
\pretocmd{\Except}{\algpx@endCodeCommand}{}{}

\ifbool{algpx@noEnd}{%
    \pretocmd{\EndTry}{\algpx@endCodeCommand[1]}{}{}%
    \pretocmd{\EndExcept}{\algpx@endCodeCommand[1]}{}{}%
}{%
    \pretocmd{\EndTry}{\algpx@endCodeCommand[0]}{}{}%
    \pretocmd{\EndExcept}{\algpx@endCodeCommand[0]}{}{}%
}%
\makeatother

\title{\modelname Eval: Evaluate Your Math LLM with your Math LLM\\--A mathematical dataset evaluation toolkit}

\author{%
  Boning Zhang\thanks{Equal contribution. This work was done when the first author's Internship at Alibaba.}~, Chengxi Li$^*$, Kai Fan$^*$\thanks{Corresponding Author}\\
  Zhejiang University, Alibaba Group \\
  \texttt{zhangbn@zju.edu.cn}, \texttt{\{xiji.lcx,k.fan\}@alibaba-inc.com} \\
}

\begin{document}

\maketitle

\begin{abstract}

Large language models (LLMs) have been explored in a variety of reasoning tasks including solving of mathematical problems. 
Each math dataset typically includes its own specially designed evaluation script, which, while suitable for its intended use, lacks generalizability across different datasets. 
Consequently, updates and adaptations to these evaluation tools tend to occur without being systematically reported, leading to inconsistencies and obstacles to fair comparison across studies. 
To bridge this gap, we introduce a comprehensive mathematical evaluation toolkit that not only utilizes a python computer algebra system (CAS) for its numerical accuracy, but also integrates an optional LLM, known for its considerable natural language processing capabilities. 
To validate the effectiveness of our toolkit, we manually annotated two distinct datasets. 
Our experiments demonstrate that the toolkit yields more robust evaluation results compared to prior works, even without an LLM. 
Furthermore, when an LLM is incorporated, there is a notable enhancement. 
The code for our method will be made available at \url{https://github.com/MARIO-Math-Reasoning/math_evaluation}.

\end{abstract}

\section{Introduction}
\label{sec:intro}

With appropriate prompts or external tools, Large language models (LLMs) have attained human parity in various tasks. 
Nonetheless, mathematical reasoning remains a formidable challenge for LLMs, necessitating a systematic evaluation to accurately compare their performance on such tasks. 
However, this field suffers from poor automatic evaluation that is neither robust nor complete enough for accurate evaluation on math problem answers, owing to their complexity and diversity. 
Recent LLMs reasoning methods (ToRA~\citep{gou2024tora} and MathChat~\citep{Wu2023AnES}) lack unified math evaluation standards even on the same dataset such as MATH~\citep{gou2024tora}. 
Such discrepancy between different evaluation scripts may not accurately reflect their true reasoning capability. 
Consequently, we are motivated to design a more comprehensive automatic evaluation toolkit that encompasses various math concepts within answers. 
Our aim is to establish a convenient and standardized evaluation framework to support future research in mathematical reasoning.

\begin{figure}
    \centering
    \includegraphics[width=0.8\textwidth]{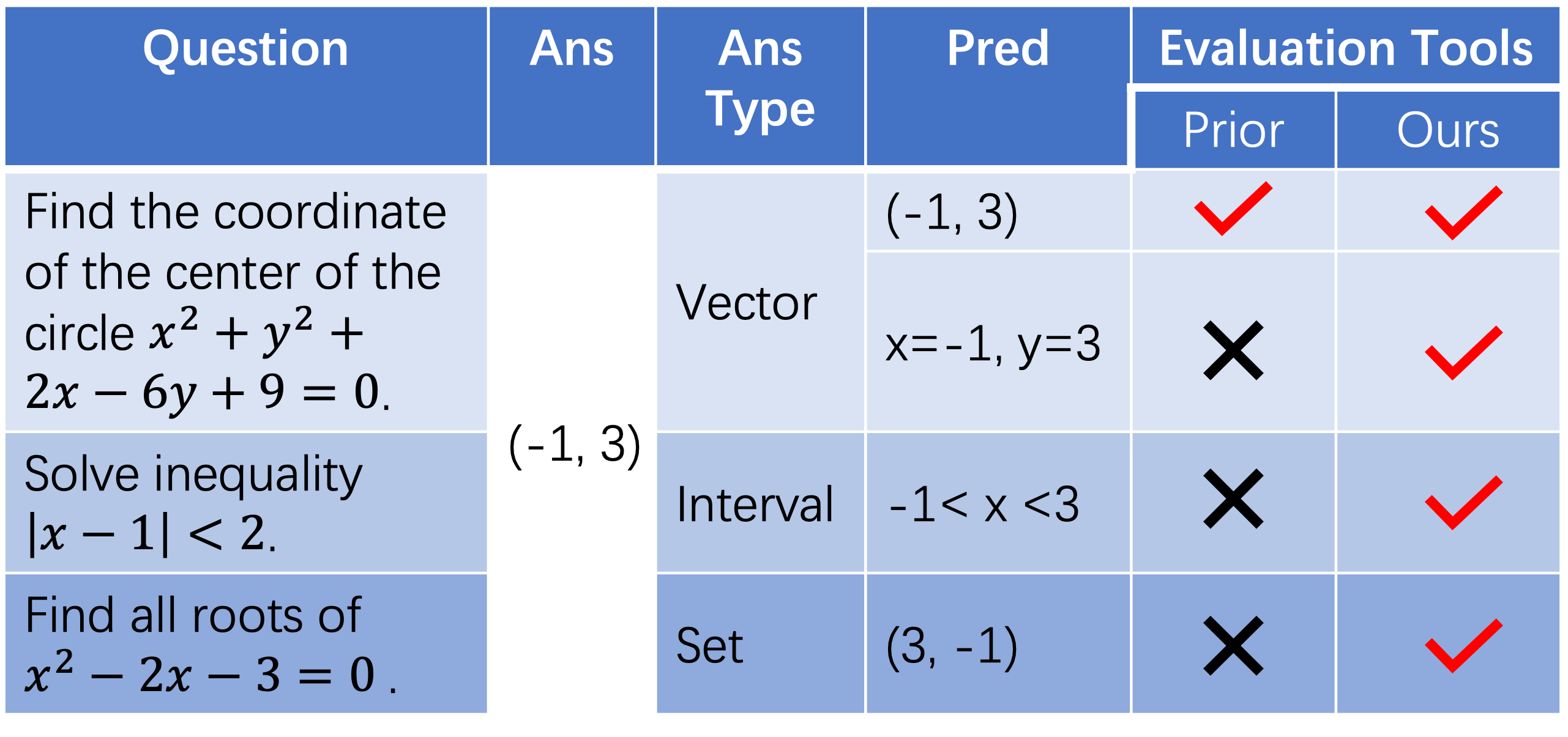}
    \caption{Most previous evaluation tools judge correctness solely based on the answer, while ours also takes into account the answer type of the question implied.}
    \label{fig:intro}
\end{figure}

Upon conducting a thorough review of existing automatic math evaluation methods, we pinpointed several key shortcomings. 
Traditionally, the assessment of mathematical answers has heavily relied on simplistic methods such as direct string comparisons or simple rules, inadequate to address complex situations. 
As illustrated in Figure~\ref{fig:intro}, identical answer expression may imply different math concepts, while different expressions may be actually equivalent under certain conditions.

\begin{table*}[ht]
\centering
\begin{tabular}{l|l}
Type & Code Friendly Definition \\
\hline
Real & A number can be used to measure a continuous one-dimensional quantity. \\
\hline
Complex & A number can be written in the form of $a + bi$, $a,b\in\mathcal{R}$ and $i$ is the imaginary unit.\\
\hline
Set & A collection of different elements which are typically mathematical objects of any kind. \\
\hline
Interval(s) & A interval is a set of real numbers that lie between two fixed endpoints without any gaps.\\
            & Intervals represent the intersection or union of several interval. \\
\hline
Vector & A collection of ordered elements which are typically mathematical objects of any kind.\\
\hline
Matrix & A rectangular array of numbers, symbols, or expressions, arranged in rows and columns. \\
\hline
Expression & A combination of symbols that is well-formed according to rules that depend on the \\
           & context, including at least one unknown variable, but no equal sign, \emph{e.g.}, $x^2 + y^2 - 1$\\
\hline
Function & A function from a set $X$ to a set $Y$ is an assignment of an element of $Y$ to each element \\
        & of $X$, \emph{e.g.}, $f(x, y) = x^2 + y^2 - 1$. \\
\hline
Equation & A equation states that two quantities are the same, in the form of $A = B$. \\
        & In our case, at least one of $A$ and $B$ should be Expression, \emph{e.g.}, $x^2 + y^2 = 1$.\\
\hline
Inequality & A relation which makes a non-equal comparison between two numbers or expressions, \\                &\emph{e.g.}, $x^2 + y^2 < 1$.\\
\hline
Others & Do not belong to above 10 types.\\
\hline
\end{tabular}
\caption{The type definitions lack complete rigor by design, with the intention to align these types with those defined in \texttt{Python} or \texttt{SymPy} as Table~\ref{tab:type_align}.}
\label{table:type_def}
\end{table*}

In light of such limitations, we propose a novel two-stage math evaluation toolkit that features an optional integration with LLM. 
The initial stage involves identifying the type of answer based on our predefined set of mathematical concepts. 
Subsequently, the second stage invokes the relevant type-specific function to evaluate the equivalence between the expected and predicted answers. 
When integrating the LLM, it has the potential to significantly enhance the precision of both the classification of answer types and the determination of answer equivalence.  
We evaluate the equivalence of answers using our toolkit in various configurations: without LLM, with LLM, and LLM-only. 
This evaluation reveals that the hybrid approach (with LLM) effectively leverages the numerical precision offered by Python packages and the natural language comprehension capabilities of the LLM. 
In summary, our key contributions are as follows:

\begin{enumerate}
    \item We construct a mathematical evaluation toolkit with an optionally integratable LLM. 
It covers a variety of mathematical reasoning answer types.

    \item We curated three mathematical answer evaluation datasets, including original questions and answers, model predicted answers, and our manually annotated answer types and equivalences. 

    \item We plan to make our datasets and evaluation toolkit publicly available to support and advance the efforts of the research community.
\end{enumerate}

\section{Main Framework}

In this section, we outline the design pattern of our mathematical evaluation toolkit. 
It mainly includes two modules. 
One is a type classification module that is utilized to ascertain the type of the expected answer. The other is an evaluation module that is crafted to assess the equivalence between the predicted response and the expected answer.

\subsection{Type Definitions}

Drawing the inspiration from the way mathematical theories are formulated by using a set of axioms, we establish our types ranging from fundamental concepts (\emph{e.g.}, Real) to complex composite structures (\emph{e.g.}, Matrix). 
The type definitions are delineated in Table~\ref{table:type_def}. 
It is important to note that these definitions are intentionally not fully rigorous; our primary aim is to ensure consistency with the types inherent within the standard \texttt{Python} language, as well as with the Python computer algebra system (CAS) package, \texttt{SymPy}. 
For instance, the type 'Set' in our framework specifically corresponds to the \texttt{set} class in \texttt{Python}, represented with curly braces (``\{...\}"), which is distinct from the concept of a mathematical set that also contains intervals. 
Table~\ref{tab:type_align} provides a comprehensive comparison between the types defined in our framework and the built-in types in \texttt{Python} and \texttt{SymPy}. 
For Real and Complex, we define the customized types, because \texttt{SymPy} considers real number as complex as well.

\begin{table}[t]
    \centering
    \begin{tabular}{c|c|c}
     Type & Package & Aligned type  \\
     \hline
     Set &  \texttt{Python}  & \texttt{set} \\
     Vector & \texttt{Python} & \texttt{list} \\
     Matrix & \texttt{SymPy} & \texttt{Matrix} \\
     Interval(s) & \texttt{SymPy} & \texttt{Interval} \\
     Expression & \texttt{SymPy} & \texttt{Expr} \\
     Function & \texttt{SymPy} & \texttt{Function} \\
     Equation & \texttt{SymPy} & \texttt{Equality} \\
     Inequality & \texttt{SymPy} & \texttt{Relational} \\
     \hline
    \end{tabular}
    \caption{Our basic design attempts to align defined types in \texttt{Python} language.}
    \label{tab:type_align}
\end{table}

\begin{algorithm}[t]
\caption{Design of our math evaluation}\label{alg:design}
\begin{algorithmic}[1]
\Require question $\mathbf{q}$, answer $\mathbf{a}$, prediction $\mathbf{p}$
\State $\mathbf{a}_{\text{type}} = \texttt{rule\_classifier}(\mathbf{a}, \mathbf{p})$
\Try
    \State $\texttt{ans} = \texttt{is\_equiv}(\mathbf{a}, \mathbf{p}, \mathbf{a}_{\text{type}})$ \Comment{False or Error may be re-evaluated later}
\EndTry
\Except
    \State $\texttt{ans} = \texttt{False}$
\EndExcept
\If{not \texttt{ans} and LLM is not None}
    \State $\mathbf{a}_{\text{type}} = \text{LLM}_{\text{type\_prompt}}(\mathbf{q}, \mathbf{a})$
    \Try
        \State $\texttt{ans} = \texttt{is\_equiv}(\mathbf{a}, \mathbf{p}, \mathbf{a}_{\text{type}})$  \Comment{Only Error will be re-evaluated}
    \EndTry
    \Except
        \State $\texttt{ans} = \text{LLM}_{\text{equiv\_prompt}}(\mathbf{q}, \mathbf{a}, \mathbf{p})$
    \EndExcept
\EndIf
\Return \texttt{ans}
\end{algorithmic}
\end{algorithm}

\subsection{Design Pattern}

Based on the definitions, we initially create a rule-based type classifier that depends solely on the expected and the predicted answers. 
For example, an answer string that starts with "\textbackslash begin{matrix}" is classified as a Matrix type. 
Subsequently, for the "Real" and "Expression" types, we develop two fundamental equivalence functions. 
For all other types, their respective equivalence functions can recursively invoke these two basic functions according to the internal data structures.

However, as Figure~\ref{fig:intro} illustrates, relying solely on the answer string to determine its type is not reliable, since the same string can represent vastly different types. 
Inspired by the remarkable natural language understanding capabilities of LLMs, we propose incorporating LLMs into the math evaluation process to eliminate the confusions highlighted in the introduction. 
With proper prompt, LLMs can analyze both the question and the answer to discern the intended answer type. 
Furthermore, they can even directly assess the equivalence between the answer and the prediction. 
To counteract the issue of hallucination, LLMs will be employed exclusively for cases where the rule-based method fails. 
The overall design is depicted in Algorithm~\ref{alg:design}.

\subsection{Datasets and Setups}

We mainly conducted experiments on two datasets, the MATH testset~\cite{hendrycks2021measuring} and GaoKao2023-Math-En (GK2023)~\cite{liao2024mario}, which contain 5,000 and 385 high school level math problems, respectively. 
We performed supervised fine-tuning on the DeepSeek-Math~\cite{deepseek-math} base model on MATH trainset and infer on the two testsets to extract the model-predicted answers. 
In addition, we downloaded the math problem solving LLM ToRA-70B~\cite{gou2024tora}. 
We then directly applied it to the GaoKao2023 dataset for inference, denoting this specific test as GK2023-ToRA.
Subsequently, we manually verified the correctness of the predicted answers for all three datasets. 

For toolkit comparison, we utilized the official repositories from \textbf{MATH}, \textbf{ToRA}, and \textbf{DeepSeek-Math}.
We assess our evaluation toolkit from three perspectives: \textbf{equivalence accuracy}, \textbf{type classification accuracy}, and \textbf{solution accuracy}.
\begin{align*}
    \texttt{equiv\_acc} &= \frac{\#\{\text{human\_eval} = \text{toolkit\_eval}\}} {\#\text{dataset}} \\
    \texttt{type\_acc} &= \frac{\#\{\text{human\_type} = \text{llm\_type}\}}{\#\text{Dataset}} \\
    \texttt{sol\_acc} &= \frac{\#\{\text{toolkit\_eval} = \texttt{True}\}}{\#\text{Dataset}}
\end{align*}

\subsection{Main Results}

The equivalence accuracy results are shown in Table~\ref{tab:equiv_acc}. 
There is a significant difference in the performance of the three open-source mathematical evaluation toolkits when assessing the same pairs of expected and predicted answers. 
Remarkably, \texttt{gpt-3.5-turbo}~\cite{openai2023gpt} appears to surpass these toolkits in terms of making correctness judgments. 
In contrast, our toolkit in the configuration of the basic design attains superior accuracy compared to \texttt{gpt-3.5-turbo}, and its performance is further enhanced when integrated with it. 

On the two testsets, our basic design can achieve about 97\% accuracy, and the incorporation of LLM technology provides an additional improvement of about 1\%. 
While this improvement may appear minor, significant accuracy enhancements are observed when integrating prior toolkits with our method outlined in Algorithm~\ref{alg:design}. These results demonstrate that LLMs can significantly amplify the effectiveness of existing tools. 
On GaoKao2023-ToRA, most of the inferred results are incorrect, \emph{i.e.}, the expected and predicted answers are quite difference. 
Despite this, all toolkits exhibited satisfactory performance, indicating that integrating the LLM may not be necessary. 
However, we can still observe that our basic design still achieves better than ToRA toolkit, which was specifically tailored for its output.

\begin{table}[t]
\centering
\begin{tabular}{l|c|c|c}
\hline
\multirow{ 2}{*}{Toolkit} & \multicolumn{3}{c}{Equivalence Accuracy}  \\ 
                \cline{2-4} 
                & MATH & GK2023 & GK2023-ToRA \\
\toprule
LLM only        & \textcolor{red}{95.03\%}   &   \textcolor{red}{93.77\%}  &  - \\ 
\midrule
MATH                           & 92.51\%    &   91.39\%  & 92.21\% \\
~~+Algorithm~\ref{alg:design}  & \textcolor{red}{95.15\%} & \textcolor{red}{93.18\%}  & -    \\
\midrule
ToRA                           & 86.08\%    &   87.24\%   &  95.06\%   \\
~~+Algorithm~\ref{alg:design}  & \textcolor{red}{96.69\%}  & \textcolor{red}{94.07\%} & -        \\
\midrule
DeepSeek-Math                  & 86.46\%   &   88.43\%   & 92.47\% \\
~~+Algorithm~\ref{alg:design}  & \textcolor{red}{96.63\%}  & \textcolor{red}{94.07\%}  & -        \\
\midrule
basic design    & 97.23\%       &   96.74\%   &  98.96\%\\
~~+ LLM type    & \textcolor{red}{97.88\%}    &   \textcolor{red}{97.33\%}   & - \\
~~~~+ LLM equiv & \textcolor{red}{98.59\%}    &   \textcolor{red}{97.03\%}   & - \\
\bottomrule
\end{tabular}%
\caption{Equivalence accuracy. LLM: \texttt{gpt-3.5-turbo}. Numbers in \textcolor{red}{red} depend on LLM.}
\label{tab:equiv_acc}
\end{table}

\begin{table}[t]
    \centering
    \begin{tabular}{c|c|c}
        \hline
       \multirow{ 2}{*}{Model} & \multicolumn{2}{c}{Accuracy on MATH testset} \\
       \cline{2-3}
        & Type & Equivalence \\
       \hline
       gpt-3.5-turbo    & 93.74\%   & 98.59\% \\
       Qwen-Max         & 93.30\%    & 97.65\%  \\
       \hline
    \end{tabular}
    \caption{Type accuracy with different LLMs}
    \label{tab:type_acc}
\end{table}

\subsection{Ablation Studies}

\noindent
\textbf{Type Accuracy} In the first study as illustrated in Table~\ref{tab:type_acc}, we specially evaluate two commercial LLMs: \texttt{gpt3.5-turbo} and Qwen-Max~\cite{bai2023qwen}. 
We choose them for two primary reasons. 
First, their cost-effective API pricing enables the handling of 5K problems. 
Because for a comprehensive assessment of the LLMs in type classification, we utilized the entire dataset with the LLMs instead of the procedure defined in Algorithm~\ref{alg:design}, activating LLM type prediction as needed. 
Second, LLMs with larger size typically exhibit enhanced abilities in natural language understanding and instruction-following, both of which are crucial for accurate type determination. 
Meanwhile, the accuracy of equivalence reported in Table~\ref{tab:type_acc} reflects the performance adhering strictly to the procedure outlined in Algorithm~\ref{alg:design}. 
In general, we can conclude that better LLMs may bring larger improvement over our basic design.

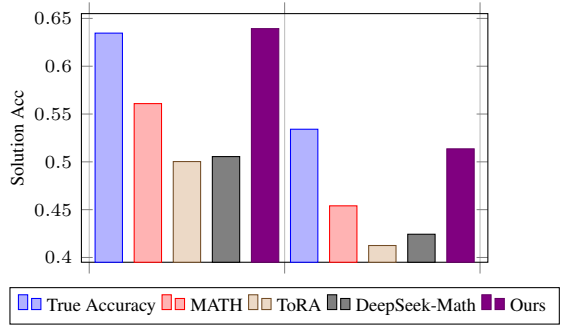
\begin{figure}[t]
\centering
\begin{tikzpicture}
\begin{axis}[
    height=0.35\textwidth, 
    width=0.5\textwidth,
    xticklabels=\empty,
    font=\scriptsize,
    ylabel=Solution Acc,
    y label style={at={(0.07,0.5)}},
    enlargelimits=0.02,
    legend style={at={(0.5,-0.1),font=\scriptsize},
    anchor=north,legend columns=-1},
    ybar interval=0.7,
    ymin=0.4, ymax=0.65,
    ytick={0.4,0.45,0.5,0.55,0.6,0.65},
]
\addplot 
	coordinates {(1, 0.6346) (2, 0.5341) (3, 0.5341) };
\addplot 
 	coordinates {(1, 0.5609) (2, 0.4540) (3, 0.4540) };
\addplot 
 	coordinates {(1, 0.5003) (2, 0.4125) (3, 0.4125) };
\addplot 
 	coordinates {(1, 0.5055) (2, 0.4243) (3, 0.4243) };
\addplot 
 	coordinates {(1, 0.6393) (2, 0.5136) (3, 0.5136) };
\legend{True Accuracy,MATH,ToRA,DeepSeek-Math,Ours}
\end{axis}
\end{tikzpicture}
    \caption{Solution accuracy with different toolkits. Left: MATH. Right: GK2023}
    \label{fig:sol_acc}
\end{figure}

\noindent
\textbf{Solution Accuracy} In the second study, depicted in Figure~\ref{fig:sol_acc}, we can observe that previous evaluation toolkits exhibit a significant discrepancy (\emph{e.g.}, over 10\% gap in the MATH dataset) from the true accuracy, thereby making it less reliable to access the actual mathematical reasoning capabilities of math LLMs. 
Someone may argue that the metric of solution accuracy may not be entirely accurate, as erroneous judgments over the correct and incorrect answers can potentially cancel each other out. 
However, we can still observe a strong correlation between equivalence accuracy and solution accuracy.  
Nonetheless, a strong correlation exists between equivalence accuracy and solution accuracy. 
For instance, in the MATH dataset and with our toolkits, higher equivalence accuracy ($>90\%$) correlates with more robust solution accuracy evaluations (less difference to the true accuracy).

\noindent
\textbf{Case Studies} We carried out three case studies using our manually annotated test sets in the Appendix~\ref{app:case_math},\ref{app:case_gaokao},\ref{app:case_math_tora}.

\section{Related Works}

In order to explore the reasoning ability of LLMs and their efficiency in understanding special symbols, researchers utilize annotated data fine-tuning, notably via `chain-of-thought' data approaches~\cite{chung2022scaling,Szegedy2020APP} and integration of external knowledge sources~\cite{thoppilan2022lamda}. 
Automated proof assistants exemplified by GPT-f \cite{Polu2020GenerativeLM} further underscores this exploration. English math word problem (MWP) corpus, such as ASDiv~\cite{Miao2020ADC}, GSM8K~\cite{cobbe2021training} and MATH~\cite{hendrycks2021measuring} are proposed to evaluate the math reasoning capabilities of LLMs in the past few years. 
However, as the most widely used dataset, reasoning experiments on MATH lack unified automatic evaluation methods. 
Our work releases a normalized math evaluation work which reaches accuracy above $95\%$ on MATH, further reflects the true capability of LLMs reasoning work more faithfully.

\section{Conclusion}

We presented a comprehensive mathematical evaluation toolkit that is able to optionally incorporate LLMs. 
This toolkit is designed to take both advantages of the CAS and LLM, and the experiments demonstrate our toolkit can promote the accuracy and consistency of math evaluation and facilitating fair cross-comparison of different studies. 

\bibliography{neurips}
\bibliographystyle{plainnat}


\section{Appendix}

\section{Case Study on our MATH}
\label{app:case_math}

In this case, there the same real number is presented in different form in the answer and predicted point. Thus the evaluation toolkit must compare the numerical equivalence between the number of coordinates to correctly assess them.
\begin{verbatim}
Example 1:
Reason: Different Form
Quesiton: Suppose $P$ is the point $(5,3)$ and $Q$ is the point $(-3,6)$. What 
is the midpoint of $\\overline{PQ}$?

Ground Truth: \\left(1,\\frac{9}{2}\\right)
prediction:   $(1.0, 4.5)$
Human: True    
Tora: False    
MATH: False
Ours: True
\end{verbatim}

In this case, same expressions are not presented in the exact same form. The evaluation toolkit may need to simplify the polynomial to the same level to evaluate the them.
\begin{verbatim}
Example 2:
Reason: Expression
Quesiton: Factor $36-4x^2$ completely.

Ground Truth: 4(3-x)(3+x)
prediction:   $-4(x - 3)(x + 3)$

Human: True    
Tora: False    
MATH: False
Ours: True
\end{verbatim}

In the following case, the answer has a followed unit string, which should be cleaned before evaluation.
\begin{verbatim}
Example 3:
Reason: Unit
Quesiton: Diana can either invest $20,\\!000$ dollars for $4$ years with a simple 
interest rate of $6\\%$ or an interest rate of $7\\%$ which compounds quarterly. 
How many more dollars, rounded to the nearest dollar, would she get with the better 
interest rate than with the worse one?

Ground Truth: 1599 \\text{ dollars}
prediction:   $1,599.00$

Human: True    
Tora: False    
MATH: False
Ours: True
\end{verbatim}

In this case, the answer is a vector, and the comparison should be applied in element-wise manner.
\begin{verbatim}
Example 4:
Reason: Elementwise comparision
Question: Convert the point $( 1, -1, -6 )$ in rectangular coordinates to cylindrical 
coordinates.  Enter your answer in the form $(r,\\theta,z),$ where $r > 0$ and 
$0 \\le \\theta < 2 \\pi.$

Ground Truth: \\left( \\sqrt{2}, \\frac{7 \\pi}{4}, -6 \\right)
Prediction:    (1.4142135623730951,5.497787143782138,-6)
Human: True    
Tora: False    
MATH: False
Ours: True
\end{verbatim}

\section{Case Study on our GAOKAO}
\label{app:case_gaokao}

In this case, the answer and the prediction are both intervals in different form. They should be converted to a interval range to be properly compared.
\begin{verbatim}
Example 1:
Reason: Interval
Quesiton: Given sets $M=\\{x|x+2\\geq 0\\},N=\\{x|x-1<0\\}$, find $M \\cap N$.

Ground Truth: x \\in [-2, 1)
prediction:   $[-2, 1)$
Human: True    
Tora: False    
MATH: False
Ours: True
\end{verbatim}

For this case, the ground truth are expressed as a set, and it means the order of the prediction does not matter.
\begin{verbatim}
Example 2:
Reason: Set
Quesiton: If the universal set is $U=\\{1,2,3,4,5\\}$, and $M=\\{1,4\\},N=\\{2,5\\}$, 
find $N \\cup \\overline{M}$.

Ground Truth: 2, 3, 5
prediction:   5, 3, 2

Human: True    
Tora: False    
MATH: False
Ours: True
\end{verbatim}
For this case, line functions are not presented in the exact same form. The evaluation toolkit may need to simplify the polynomial on the right side to the same level, then evaluate the function equation.
\begin{verbatim}
Example 3:
Reason: Expression
Quesiton: Find the tangent line to the function $y=\\frac{e^{x}}{x+1}$ at point
$\\left(1,\\frac{e}{2}\\right)$.

Ground Truth: y=\\frac{e}{4}x+\\frac{e}{4}
prediction:   $f(x) = \\frac{e(x + 1)}{4}$

Human: True    
Tora: False    
MATH: False
Ours: True
\end{verbatim}

\section{Case Study on MATH of ToRA}
\label{app:case_math_tora}
In this instance, a set of values should be answered to the question. So the order of numbers separated by commas does not matter.
\begin{verbatim}
Example 1:
Reason: Set
Quesiton: A line segment of length $5$ has one endpoint at $(1, 2)$ and 
the other endpoint at $(4, b)$. Find all possible values of $b$, separated by commas.

Ground Truth: 6,-2
prediction:   -2,6
Human: True    
Tora: False    
MATH: False
Ours: True
\end{verbatim}
\end{document}